\documentclass[conference]{IEEEtran}
\IEEEoverridecommandlockouts
\usepackage{colortbl}
\usepackage{color}
\usepackage{xcolor}
\usepackage{cite}
\usepackage{amsmath,amssymb,amsfonts}
\usepackage{algorithmic}
\usepackage{graphicx}
\usepackage{textcomp}
\usepackage{hyperref}
\usepackage{tikz,pgf,pgfplots}
\usetikzlibrary{patterns}
\usetikzlibrary{shapes,arrows}
\usetikzlibrary{plotmarks}
\usetikzlibrary{external} 
\usetikzlibrary{decorations.pathmorphing}
\usetikzlibrary{arrows.meta}
\usetikzlibrary{positioning,fit}
\usepackage{subcaption}

\definecolor{verylightgray}{gray}{.75}
\definecolor{veryverylightgray}{gray}{.85}

\usepackage[numbers]{natbib}
\renewcommand{\cite}[1]{\citep{#1}}

\usepackage{sty/slashbox}
\usepackage{booktabs}

\pgfplotsset{compat=newest}
\usepackage{tikzscale}     

\def\BibTeX{{\rm B\kern-.05em{\sc i\kern-.025em b}\kern-.08em
    T\kern-.1667em\lower.7ex\hbox{E}\kern-.125emX}}

\newcommand{ \MMCneg  }{$\text{MMC}_{\mathrm{inc}} $}

\newcommand{\arrACC}{Acc $\uparrow$}

\newcommand{\arrMMCneg}{$\text{MMC}_{\mathrm{inc}}$ $\downarrow$}

\newcommand{\arrECE}{ECE $\downarrow$}

\begin{document}

\title{Improving Uncertainty of Deep Learning-based Object Classification on Radar Spectra using Label Smoothing}

\author{
\vspace{-0.1cm}
\IEEEauthorblockN{Kanil Patel$^{1,2}$,
William Beluch$^{1}$,
Kilian Rambach$^{1}$, 
Michael Pfeiffer$^{1}$,
Bin Yang$^{2}$
} \\
\vspace{-0.2cm}
\IEEEauthorblockA{
$^{1}$Bosch Center for Artificial Intelligence, Renningen, Germany \\
$^{2}$Institute of Signal Processing and System Theory, University of Stuttgart, Stuttgart, Germany \\
}}

\maketitle

\begin{abstract}
    Object type classification for automotive radar has greatly improved with recent deep learning (DL) solutions, however these developments have mostly focused on the classification accuracy.
    Before employing DL solutions in safety-critical applications, such as automated driving, an indispensable prerequisite is the accurate quantification of the classifiers' reliability.
    Unfortunately, DL classifiers are characterized as black-box systems which output severely over-confident predictions, leading downstream decision-making systems to false conclusions with possibly catastrophic consequences.
    We find that deep radar classifiers maintain high-confidences for ambiguous, difficult samples, e.g. small objects measured at large distances, under domain shift and signal corruptions, regardless of the correctness of the predictions.
    The focus of this article is to learn deep radar spectra classifiers which offer robust real-time uncertainty estimates using label smoothing during training.
    Label smoothing is a technique of refining, or \emph{softening}, the hard labels typically available in classification datasets.
    In this article, we exploit radar-specific know-how to define soft labels which encourage the classifiers to learn to output high-quality calibrated uncertainty estimates, thereby partially resolving the problem of over-confidence.
    Our investigations show how simple radar knowledge can easily be combined with complex data-driven learning algorithms to yield safe automotive radar perception.

\end{abstract}

\section{Introduction}

As an important component of an automated driving system, radar sensors play a crucial role in the safe and robust perception of the environment.
Recently, deep learning (DL) based solutions have shown prodigious performance in accurately classifying the object type from radar spectra in the presence of labeled data~\cite{patel_radar19,radarspectra_augmentations_sheeny2020radio}.
However, developments have mostly focused on improving the \emph{generalization accuracy} instead of the robustness, reliability or uncertainty of the predictions.
As a result, softmax predictions from such high capacity models tend to be highly accurate, but poor representatives of the predictive uncertainty~\cite{pmlr-v70-guo17a, patel_radar21}.
Classifiers exhibiting such characteristics, albeit most often accurate, tend to be over-confident and have limited use in practice, as decision-making systems fail to distinguish between incorrect over-confident predictions and correct high-confident predictions~\cite{patel_radar21}.

Among multiple reasons behind this notorious over-confidence of DL classifiers, one fundamental reason emanates from the dataset used during training, specifically the labels~\cite{epssmoothing, patel2019onmanifold}.
The labels of supervised datasets are typically one-hot label vectors (i.e. a binary label for each class) with a single ground truth class label (i.e. label vectors summing to $1$).
These \emph{hard} labels are the most accurate representation of the true classification of an object.
However, given their binary nature they provide no information of the uncertainty inherent in the data and induce an unwanted over-confidence bias in the learned predictions.
Alternatively, \emph{soft} labels~\cite{epssmoothing} are non-binary categorical distributions which better quantify the ambiguity present in the data.
In this article, we identify that the over-confidence in deep radar classifiers, which emanates from using hard labels, can be fixed using soft labels~\cite{whenlabelsmoothinghelps,HintonDistilation,OnMixupTrainThul, patel2019onmanifold} and propose two novel heuristics to compute sample-specific smoothing factors to refine the hard labels.

\begin{figure*}[htb]
\begin{subfigure}{0.32\textwidth}
    \includegraphics[width=1.0\textwidth, height=3.8cm]{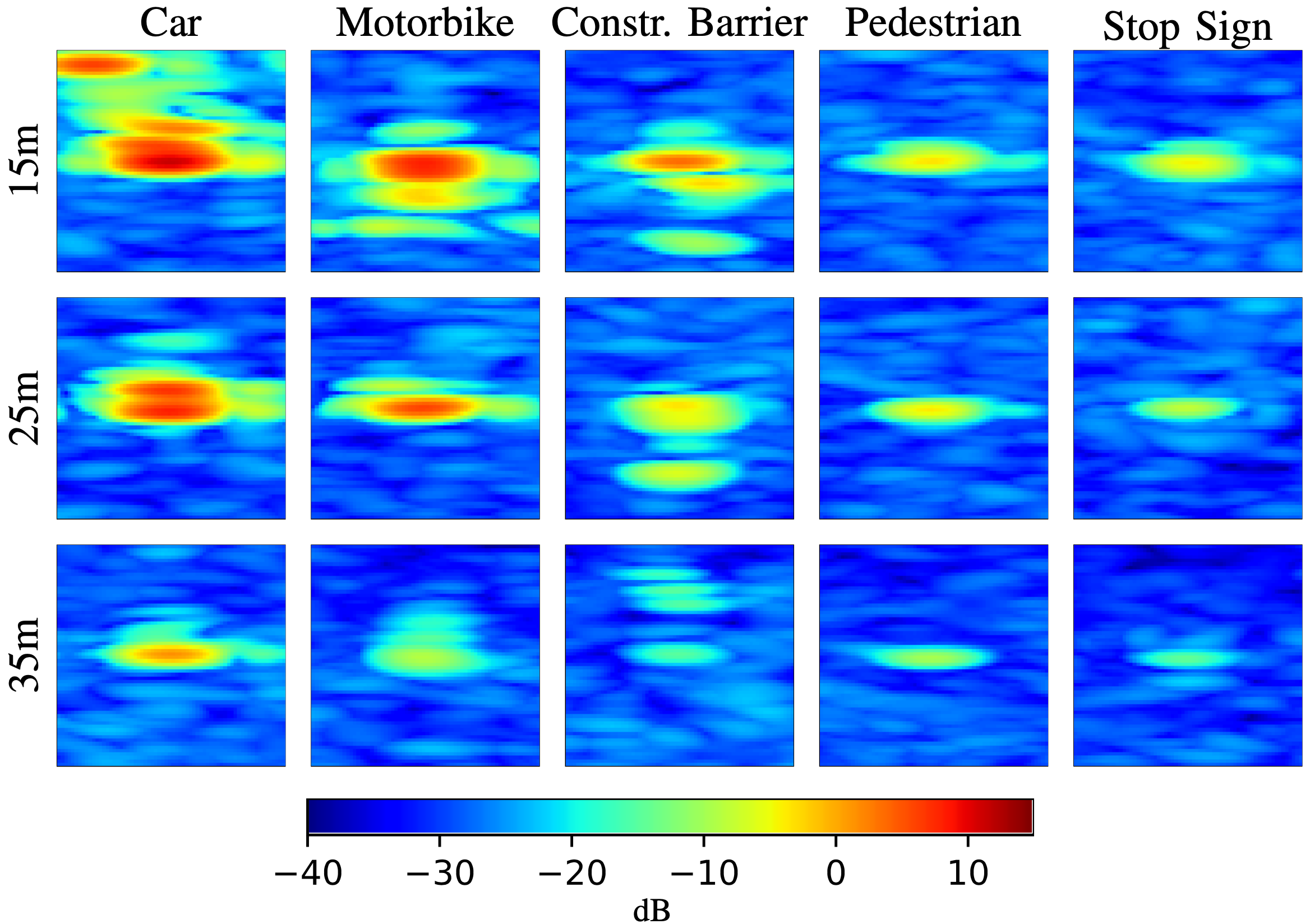}
    \caption{}
    \label{fig:spectra_roi_samples}
\end{subfigure}\hfill
\begin{subfigure}{0.32\textwidth}
    \includegraphics[width=1.0\textwidth]{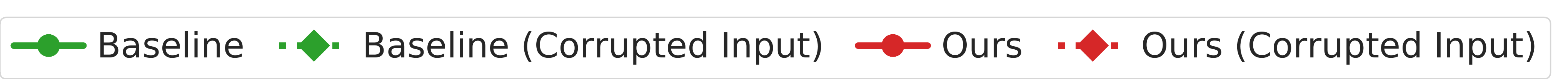}
    \includegraphics[width=1.0\textwidth, height=3.5cm]{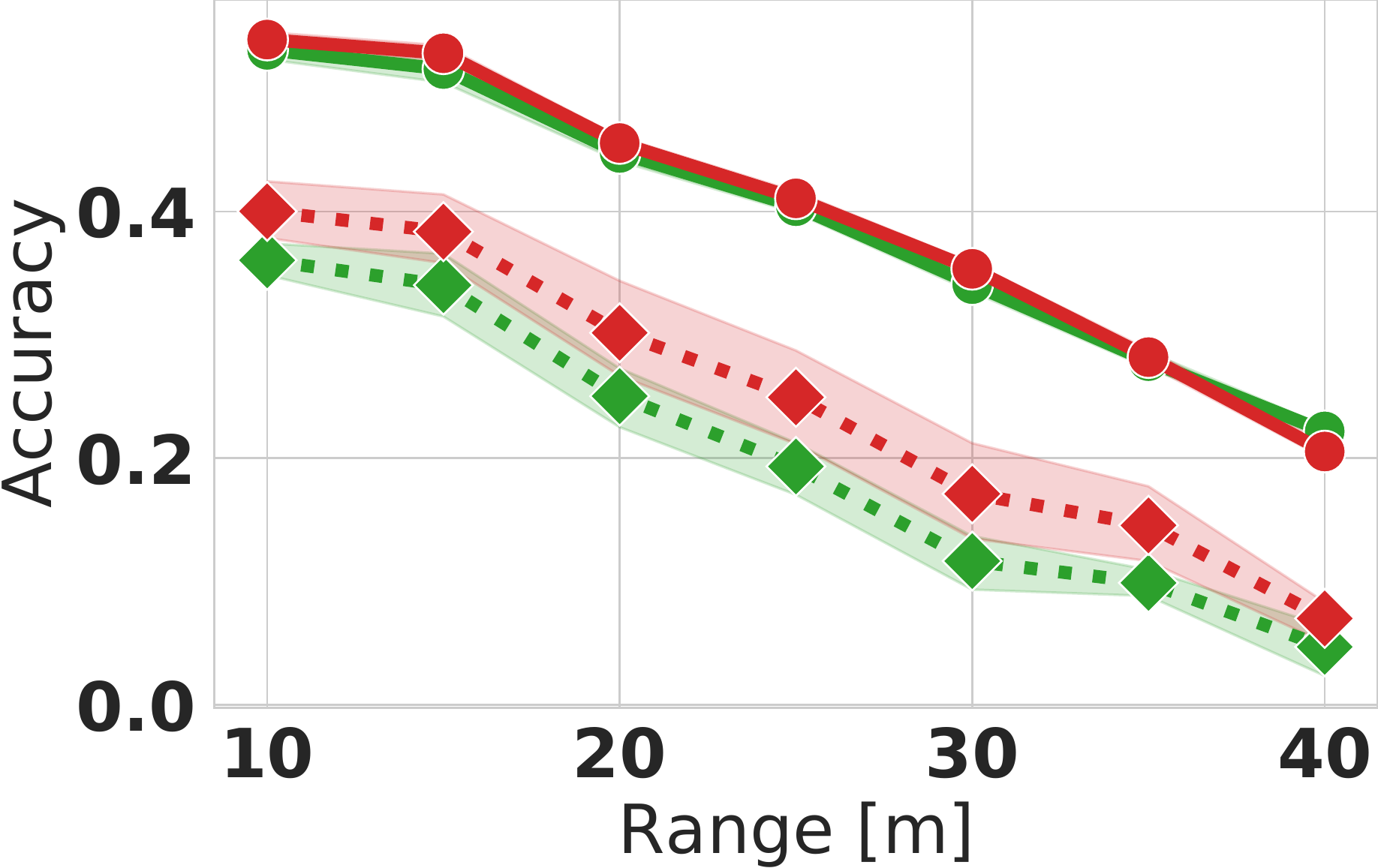}
    \caption{}
    \label{fig:teaser_baseline_vs_ours}
\end{subfigure}
\hfill
\begin{subfigure}{0.32\textwidth}
    \includegraphics[width=1.0\textwidth]{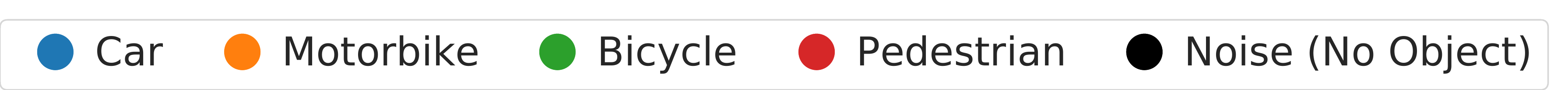}
    \includegraphics[width=1.0\textwidth, height=3.5cm]{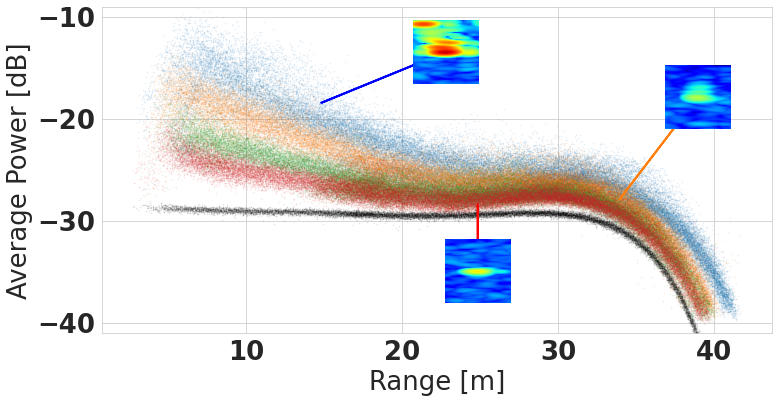}
    \caption{}
    \label{fig:scatter_rcs_vs_range_plot}
\end{subfigure}

    \caption{
        \textbf{(a)} 
            \textbf{Range-Azimuth spectra samples} of $5$ objects each measured at $3$ different distances.
            We highlight that the number and amplitudes of the peaks reduce for each object when measured from farther away and for a given range the total received power differs across each object.
            As expected, objects with large metallic components (e.g. car and motorbike) are much more reflective than pedestrians and stop signs which only have few small or no metallic components.
        \textbf{(b)} 
            \textbf{Classification accuracy} of the Baseline and our soft-label classifier ("Ours") on the test set and the corrupted test set (speckle noise at severity $1$) at varying ranges.
            The performance significantly degrades at larger distances with better performance seen by our label smoothing method.
            More information can be found in Sec.~\ref{sec:p5_performance_over_range}.
        \textbf{(c)}
            \textbf{Average received power} scatter plot of all ROI spectra and the ground truth range.
            We again observe the received power degrades over range and that overall some objects reflect more power than others. 
            We also plot the average power of the Noise, which are ROI spectra where no objects were present.
    }
\end{figure*}

\textbf{Sources of uncertainty in radar spectra}:
In radar spectra classification, even in the absence of corruptions or sensor malfunctions, there exists ambiguity in the spectra that cannot be represented with hard labels.
For example, measuring the same object from $2$ different distances will share the same hard label even though the measurement from farther away could be harder to classifiy.
According to the radar range equation, the power measured by the receiving antennas depends, among other factors, on the amount of transmitted power, the range, and reflecting characteristics of the objects.
As the power transmitted to all objects in the field-of-view remains roughly uniform, the received power is some inaccessible complex function of the range and reflecting characteristics of the object.
The received power, measured in the spectra, is used by deep learning classifiers to approximate this complex function by finding features which lead to the object's accurate classification. 
As the received power decreases and less class-specific information is available in the spectra, the object class becomes ambiguous, ultimately making it harder to classify as the classifiers rely on much less information.
This ambiguity can increase for small, low-reflective objects or objects at large distances which reflect relatively fewer or lower power peaks.
Fig.~\ref{fig:spectra_roi_samples} visualizes parts of the range-azimuth spectra for various objects, called region-of-interest (ROI) samples, measured at varying ranges which show how both these effects control the amount of received power.

\textbf{Soft labels in the radar domain}:
Obtaining soft-labels which accurately quantify the uncertainty of a sample can often be arduous (i.e. humans are not good at accurately quantifying uncertainty), expensive (i.e. obtaining soft labels by aggregating multiple annotator hard labels for large datasets is costly) or, at times, even impossible (i.e. human annotators are not always involved in the labeling process) to obtain.
For radar spectra, it is even harder as humans are not able to easily identify objects solely based on their spectra.
For smaller distant objects such as the stop sign at $35$m, it is clearly visible in Fig.~\ref{fig:spectra_roi_samples} that there is only a single small peak close to the noise floor.
For such samples, the radar classifiers are tasked with the challenge of providing a prediction based on this single relatively small peak.
The smaller and fewer the peaks become (e.g. stop sign at even larger distances), the harder the classification tasks becomes as a similar small peak can also be received from other small objects (e.g. pedestrian) or large objects at large distances (e.g. car at $80$m).
The difficulty in predicting distant samples can be observed in Fig.~\ref{fig:teaser_baseline_vs_ours}, where we evaluate the accuracy of the Baseline (i.e. trained with hard labels) and Ours (i.e. using soft labels as presented in Sec.~\ref{sec:p5_methodology}) on the test set at varying distances.
Additionally, we evaluate the test set corrupted with speckle noise at severity $1$ (i.e. the dotted curves).
We observe that overall the performance degrades at larger distances as the ROI samples become increasingly ambiguous with range.

\textbf{Exploiting radar know-how for label smoothing}: 
Using this observation, we exploit the range and received power of the spectra to refine the hard labels to better reflect the uncertainty or difficulty in predicting the spectra.
We note that there are multiple ways to approximate the uncertainty associated with a radar spectra, though we find that using these simple measures are sufficient to significantly improve the calibration performance of the classifiers.
Using other measures is a part of future works.
In Fig.~\ref{fig:scatter_rcs_vs_range_plot}, we plot the average received power of each region-of-interest (ROI) spectra and plot it against the range for multiple object classes for the entire dataset.
It is seen that the received power greatly decreases when objects are farther away, due to the loss of power with distance, and larger objects have significantly stronger reflections.
We leverage this information as a proxy to the uncertainty of the spectra to smooth the hard labels.

\section{Related Work}

Numerous solutions have been proposed for radar classification using point-cloud and spectra representations in the context of automated driving.
The class of algorithms operating on point-cloud radar reflections, first require applying a statistical detection algorithm (e.g. CFAR~\cite{rohlingCFAR}) followed by a clustering algorithm which attempts to group reflections from the same object.
Given these clusters, classification algorithms can be applied on either the extracted hand-crafted features~\cite{zhao2020_handcraftedfeatures,2019ghost_handcraftfeats,dube2014detection} or learned features~\cite{ulrich2020deepreflecs}.
The clustering step is avoided in~\cite{kraus2020GhostImagesRadar,feng2019point,schumann2018} by directly using deep learning to yield a semantic segmentation of the point cloud.
The point-cloud can also be exploited to create occupancy grids which can directly be used for classification~\cite{ensembleLombacher2017,staticObjectClassificationDL-lombacher2016}.

The alternative spectra representation is a result of applying a multi-dimensional FFT on the raw radar signals and produces image-like maps of the environments measuring the range, azimuth and Doppler velocity.
Most learning-based tasks exploit micro-Doppler signatures of moving objects~\cite{palffy2020cnn,major2019vehicle,perez2018single,angelov2018practical}, but range-azimuth maps have also been used for various tasks~\cite{rodnet_radar_obj_det}, especially for static scenes where the micro-Doppler signatures are small or non-existant~\cite{patel_radar19, radarspectra_augmentations_sheeny2020radio}.

In deep learning, an effective technique for uncertainty estimation involves computing them from multiple forward passes via sampling or ensembles.
Albeit powerful, this family of methods produce unwanted latencies which are inapt for automotive applications.
Computationally efficient techniques which are better suited for real-time applications include sampling-free uncertainty estimation \cite{Postels_2019_ICCV} and data augmentation \cite{patel2019onmanifold, OnMixupTrainThul} which are employed during training.
Alternatively, post-hoc calibration techniques~\cite{pmlr-v70-guo17a, patel2020imax} improve uncertainty estimates without retraining and were applied to radar spectra in~\cite{patel_radar21}.
This work is orthogonal to the post-hoc calibration work presented in~\cite{patel_radar21} and can be combined by applying~\cite{patel_radar21} after applying the techniques presented here.

Since the pioneering work of \cite{HintonDistilation} and \cite{epssmoothing}, introduced to improve the generation accuracy, recently another family of methods include using soft labels during training for improving the uncertainty estimation of neural networks~\cite{OnMixupTrainThul,ReLUMHein,patel2019onmanifold}.
These methods involve refining the hard labels available in the supervised datasets to quantify the inherent uncertainty in the data.
During the training these soft labels allow the classifier to reflect ambiguous and uncertain situations.

In this paper, we introduce two variants of a novel soft label training procedure which produces sample-specific label smoothing factors using available information such as the range of the object or the amount of received power in the spectra.
We note that this is the first work applying soft labels for radar spectra classification.

\section{Methodology}
\label{sec:p5_methodology}
We formally introduce soft labels, the concept of label smoothing~\cite{epssmoothing} and the two novel techniques, called \textbf{R-smoothing} and \textbf{P-smoothing}, presented in this work.

\subsection{Hard labels vs. soft labels}
\label{sec:hard_vs_soft_labels}
In a classification setting, the ground truth data distribution $P_* (x,y)$ is unknown and approximated by a finite training dataset, 
$D={\{x_i,y_i \}}_{i=1}^N$ (i.e. empirical distribution $P_d (x,y)=1/N \sum_{(i=1)}^N \delta(x=x_i,y=y_i))$.
The empirical distribution is formed by assembling delta functions located on each example~\cite{Zhang2018mixupBE}, with further improvements possible by replacing them with some density estimate of the vicinity~\cite{vicianalriskmin}.
Using expert domain knowledge, the vicinity around the data can be defined for further sample and label generation.
Even though sample generation, known as data augmentation, is commonly performed, the labels $y$ are often left unchanged.
For a $C$-class classification problem, these labels $y$ are hard estimates (i.e. one-hot encodings: $y \in \{y': y' \in \{0,1\}^C, 1^T y'=1\}$) of the true conditional distribution $P_* (y | x)$.
Training with hard labels has the adverse effect of producing over-confident classifiers as they are solely trained on zero-entropy labels~\cite{patel2019onmanifold}, therefore penalizing any sign of uncertainty reflected in the predictions.
These predictions also tend to be highly mis-calibrated~\cite{pmlr-v70-guo17a,patel2020imax} as the training loss tends to encourage fully confident predictions, regardless of their correctness.

An alternative, which can better reflect the true confidences for each class, is obtained by using (non-binary) soft labels (i.e. $y \in \{y': y' \in {[0,1]}^C, 1^T y'=1\}$).
Soft labels have the capacity to discourage over-confidence by assigning lower confidences for ambiguous and uncertain samples, though at the same time maintaining the true class label information.
An open problem which remains in the literature is the acquisition of these soft labels as it is often difficult to quantify uncertainty.
One simple, yet effective, approach is to uniformly smooth all labels in the dataset but does not consider any class or input specific information.
In this work, we aim to introduce two techniques for acquiring soft labels for radar spectra to improve the uncertainty of the trained classifiers.

\subsection{Label smoothing}
In the pioneering work of label smoothing~\cite{epssmoothing}, the authors use a fixed value $\epsilon$ to smooth \emph{all} hard labels, called $\epsilon$-smoothing.
$\epsilon$-smoothing can be seen as a mixture between the hard labels $y$ and the class prior distribution over labels $\upsilon$, with weights $1 - \epsilon$ and $\epsilon$, respectively.
The resulting labels regularize the training procedure and partially address the issue of over-confidence for smoothing values even as small as $0.01-0.1$.
However, this approach is agnostic to the ambiguity or uncertainty inherently present in each sample.
More formally, the label $y$ is smoothed to 
\begin{equation}
    \tilde{y} = (1 - \epsilon)y + \epsilon\upsilon, 
    \label{eq:labelsmoothing}
\end{equation}    
where the value of $\epsilon$ determines the amount of smoothing.
Compared to assigning $\epsilon$ to a small \emph{fixed} value (with best validation set performance achieved using $\epsilon=0.1$~\cite{epssmoothing}), we propose to use an adaptive $\epsilon$ for each sample for further calibration performance gains.

\subsection{Estimating soft labels for radar spectra: R-smoothing and P-smoothing}
Building on the label smoothing technique, we propose \textbf{R-smoothing}, which uses the range $R$ of the object to determine the value of $\epsilon$,
\begin{equation}
    \epsilon_{R} = 1 - e ^{  -\alpha \frac{R - r_{\text{min}}}{r_{\text{max}} - r_{\text{min}}}  },
\end{equation}    
where $\alpha > 0$ controls the smoothing factor, $r_{\text{min}}$ and $r_{\text{max}}$ are the minimum and maximum range values for training samples and are used to normalize the range to $[0,1]$.
This smoothing technique will assign lower confidences (i.e. larger $\epsilon$) for objects farther away, which are known to pose a greater challenge in determining its classification (see Fig.~\ref{fig:teaser_baseline_vs_ours}) due to the reduced information present in the spectra (see Fig.~\ref{fig:scatter_rcs_vs_range_plot}).
However, we note that this smoothing is agnostic to the object class and smoothes each objects' label uniformly, without leveraging any input or object specific information.

To address this, we additionally propose \textbf{P-smoothing}, which uses the average received power of the spectra sample $\pi(x) = \frac{1}{P} \sum_{p=1}^{P} x_p$ for the \emph{linear scale} input $x$ with $P$ pixels to determine the smoothing factor $\epsilon$,
\begin{equation}
    \epsilon_{\pi} = 1 - e ^{  -\alpha \left( 1- \frac{\pi(x) - \pi_{\text{min}}}{ \pi_{\text{max}} - \pi_{\text{min}}  }  \right) },
\end{equation}    
where $\alpha > 0$ controls the smoothing factor, $\pi_{\text{min}}$ and $\pi_{\text{max}}$ are the minimum and maximum average power for spectra samples in the training dataset.
In order to ensure that the soft labels still assign more than $50\%$ to the ground truth class, we bound the hyper-parameter $\alpha$ by $0 < \alpha < -\log_e{0.5}$.
As the spectra of highly reflective objects will contain more received power and signal information, their labels will have smaller $\epsilon$, indicating lower uncertainty.
The distribution of the average power measured across range can be seen in Fig.~\ref{fig:scatter_rcs_vs_range_plot}.
The average power has been used in \cite{tristanphd} as a measure to filter out noisy samples to clean the training dataset.
However, instead of using this measure as a filter to remove (noisy) samples, we propose to use it to determine the degree to which the hard labels should be smoothed. 
This allows the classifier to still learn from these noisy samples, but is encouraged to assign them significantly lower confidences.

\section{Experimental setup}
\textbf{Dataset}:
We use the same measurement framework and radar sensor as described in~\cite{patel_radar19} to measure, pre-process, label, and extract the regions-of-interests (ROIs) from radar spectra.
We also use the two static scene datasets introduced in~\cite{patel_radar21} (now including objects up to $43$m in range instead of the previous $30$m) which consist of the following seven objects: car, construction barrier, motorbike, baby carriage, bicycle, pedestrian, and stop sign.
These objects are measured by an ego-vehicle driving through the static scene in different driving patterns.
Using a similar setup, we additionally create a new test set, Env3-Test, which consists of a novel scene and driving pattern.
Env3-Test only consists of driving patterns which drive straight towards the objects in the scene.
All datasets differ in the measured scenes, driving patterns, object instances and viewing angles.
Each dataset consists of $414 281$, $49 726$ and $355 544$ data samples for Env1, Env2-Test and Env3-Test, respectively, where only part of Env1 is used during the training and validation phase.
Env2-Test and Env3-Test are used for an unseen evaluation of the performance.

\textbf{Training}:
We use the same architecture and training procedures as described in~\cite{patel_radar21}.
We select the classifier with the best accuracy only using a split of Env1 and evaluate the performance on the unseen test datasets: Env2-Test and Env3-Test.
We train $10$ independent classifiers and report their mean and standard deviations for all experiments.
We used a held-out validation set to determine the hyper-parameters of each method, yielding the best accuracy performance using $\epsilon=0.1$ for $\epsilon$-smoothing and $\alpha=0.5$ for R- and P-smoothing.

\textbf{Uncertainty Metrics}: 
The quality of the confidence estimates can be evaluated using the Expected Calibration Error (ECE)~\cite{pmlr-v70-guo17a} and Mean Maximal Confidence (MMC).
The ECE gives an indication of the correctness of the confidence estimates.
This is calculated by grouping predicted samples into discrete bins $b$.
More formally, ECE is computed as:
\begin{equation}
\mathrm{ECE} = \sum_{r=1}^{N_{B}} \frac{N_{b}}{N} | \mathrm{accuracy}(b) - \mathrm{confidence}(b)| ,
\end{equation}
for $N$ samples, $N_B$ bins, $N_b$ samples in bin $b$, and the calibration range $b$ is defined by the $[N/N_B]^{\mathrm{th}}$ index of the sorted predicted confidences. 

The MMC is the mean confidence across a set of samples, with the confidence being the maximum predicted class probability.
We examine the MMC of correct vs. incorrect classifications, for which we wish the former set to have a high MMC, and the latter sets to have a low MMC.  
More details on the ECE and MMC can be found in~\cite{patel_radar21}.

\section{Uncertainty calibration}
\label{sec:p5_exps_uncertainty_cal}
\begin{figure}[!t]
\centering
\begin{subfigure}{0.75\columnwidth}
  \centering
  \includegraphics[width=1.0\textwidth, clip]{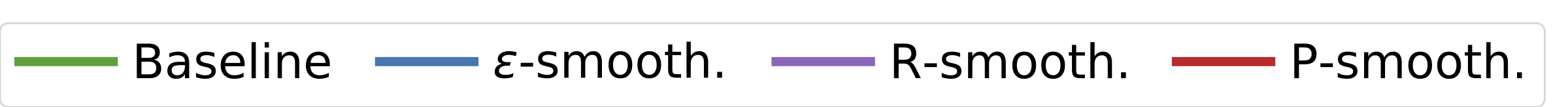}
  \label{legend}
\end{subfigure}	
    \vspace{-0.3cm}
\\
    \begin{subfigure}[t]{0.49\columnwidth}
		\includegraphics[width=0.95\textwidth]{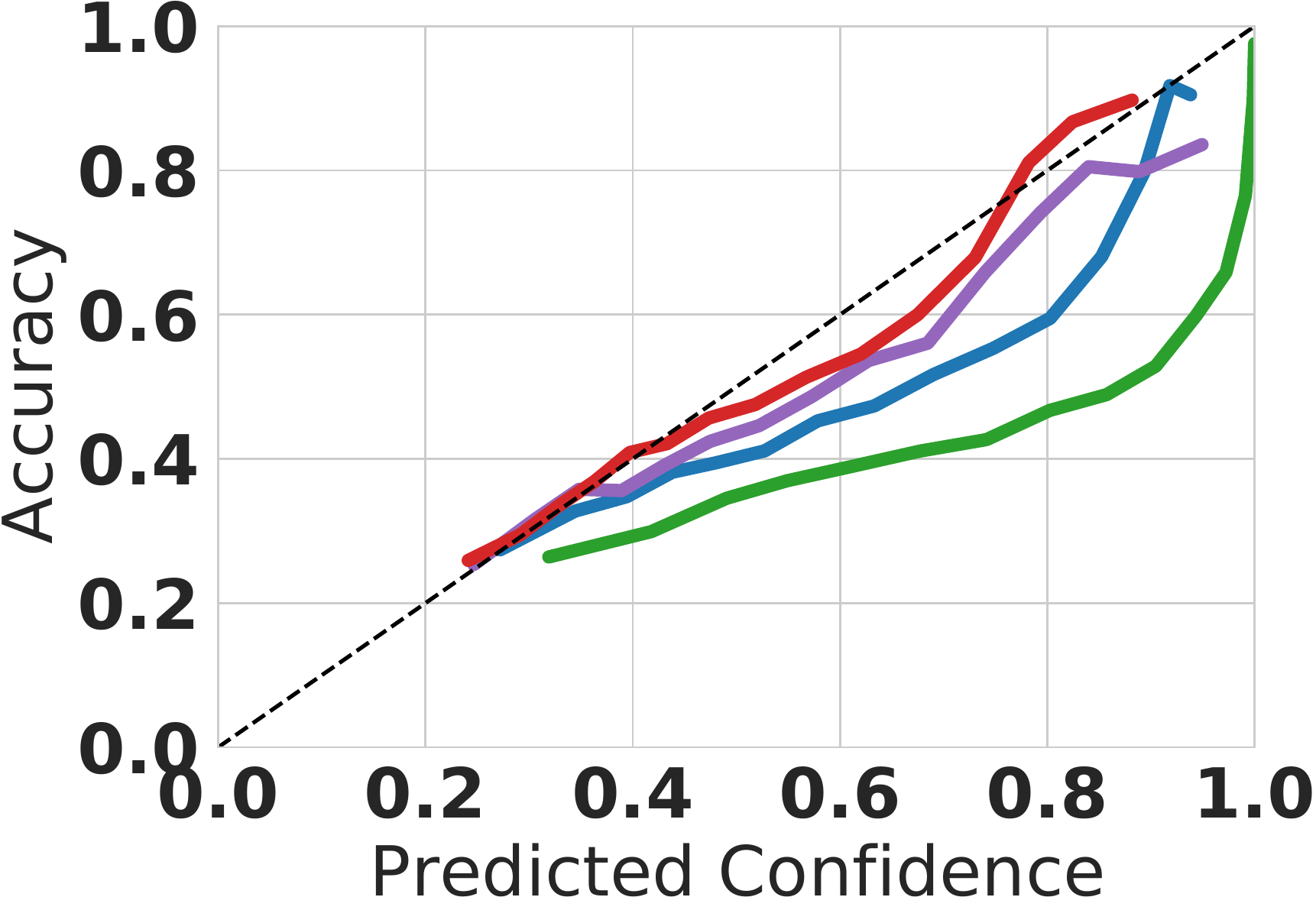}
		\caption{Env2-Test}
	\end{subfigure}
	\hfill
	\begin{subfigure}[t]{0.49\columnwidth}
		\includegraphics[width=0.95\textwidth]{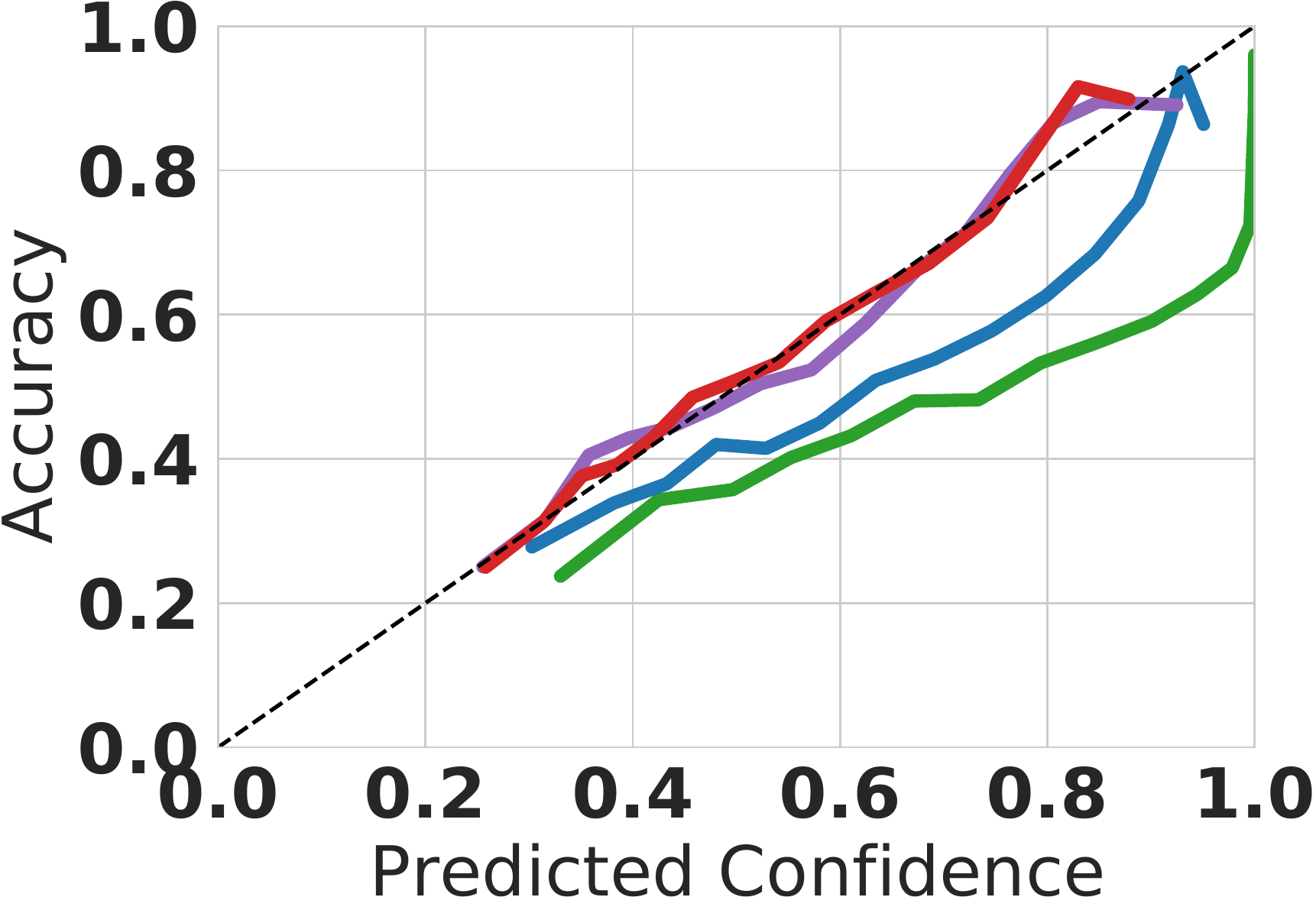}
		\caption{Env3-Test}
	\end{subfigure}
    \vspace{-0.1cm}
    \caption{Reliability diagrams of the Baseline and label smoothing methods for (a) Env2-Test and (b) Env3-Test. 
    Using soft labels instead of the hard labels (i.e. Baseline) greatly improves the confidence calibration.
Among these, overall P-smoothing shows the best calibration performance.
}
\label{fig:cal_curves_soft_labels}
\end{figure}

In Fig.~\ref{fig:cal_curves_soft_labels}, we plot the reliability diagrams of all methods evaluated on the two test sets.
A classifier has perfect calibration when its predictive confidences match the accuracy (i.e. the black dashed diagonal line), and is over-confident (under-confident) when below (above) this line.
The severe over-confidence of the Baseline can be seen by the skewness of the curve to the right and the largest distance to the ideal calibration curve.
We observe that all label smoothing methods improve the calibration performance and that R-smoothing and P-smoothing perform best. 
As this is merely a qualitative evaluation to visually compare the calibration performance, in the next section we provide more quantitative evaluations.

\section{Quantitative evaluations of all metrics}
\label{sec:p5_exps_big_table}
In Tab.~\ref{tab:main_table_test_sets}, we quantitatively evaluate the Accuracy, ECE and MMC on the test sets, Env2-Test and Env3-Test. 
We observe that all label smoothing methods improve the Baseline across all metrics, with the best performance seen with R-smoothing and P-smoothing.
Even though the soft labels used by R-/P-smoothing were designed to improve the calibration performance and address the issues of over-confidence, we also observe an accuracy improvement. 
In addition to significantly improving the calibration performance, the classifiers are also less over-confident on the incorrectly classified samples (i.e. lower \MMCneg values).
This shows that the classifiers generalize better when using the modified soft labels instead of the ground truth hard labels.
\begin{table}[!htp]
\footnotesize
\centering
\caption{
    Quantitative evaluation of the Baseline and label smoothing methods on Accuracy, ECE, and \MMCneg.
}
\label{tab:main_table_test_sets}
\begin{tabular}{l|ccc}
\toprule
\rowcolor{verylightgray}				        
      Method & \arrACC & \arrECE & \arrMMCneg \\ 
\midrule
\rowcolor{veryverylightgray}
				   &    \multicolumn{3}{c}{Env2-Test}  \\
    Baseline            &  52.50 $\pm$ 0.32 &  0.227 $\pm$ 0.03 &  0.666 $\pm$ 0.03 \\
    $\epsilon$-smooth.  &  53.05 $\pm$ 0.34 &  0.108 $\pm$ 0.01 &  0.553 $\pm$ 0.01 \\
    R-smooth.           &  53.15 $\pm$ 0.48 &  0.048 $\pm$ 0.01 &  0.486 $\pm$ 0.01 \\
    P-smooth.         &  \textbf{53.50} $\pm$ 0.61 &  \textbf{0.036} $\pm$ 0.00 &  \textbf{0.452} $\pm$ 0.01 \\

\rowcolor{veryverylightgray}
                   &    \multicolumn{3}{c}{Env3-Test}  \\
    Baseline           &  57.05 $\pm$ 0.27 &  0.191 $\pm$ 0.02 &  0.671 $\pm$ 0.02 \\
    $\epsilon$-smooth. &  57.35 $\pm$ 0.21 &  0.102 $\pm$ 0.01 &  0.579 $\pm$ 0.01 \\
    R-smooth.          &  \textbf{57.78} $\pm$ 0.36 &  \textbf{0.033} $\pm$ 0.00 & 0.478 $\pm$ 0.01 \\
    P-smooth.        &  57.40 $\pm$ 0.40 &  \textbf{0.033} $\pm$ 0.00 &  \textbf{0.467} $\pm$ 0.01 \\
\end{tabular}
\end{table}

\section{Study of the performance over object range} 
\label{sec:p5_performance_over_range}
\begin{figure} [!htp]
\centering
\begin{subfigure}{0.95\columnwidth}
  \centering
  \includegraphics[width=0.95\textwidth, clip]{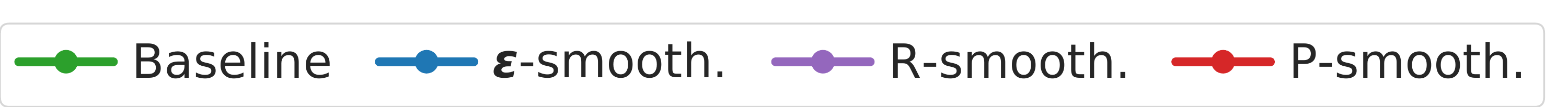}
\end{subfigure}
\\
\begin{subfigure}{0.48\columnwidth}
  \centering
  \includegraphics[width=1.0\textwidth, clip]{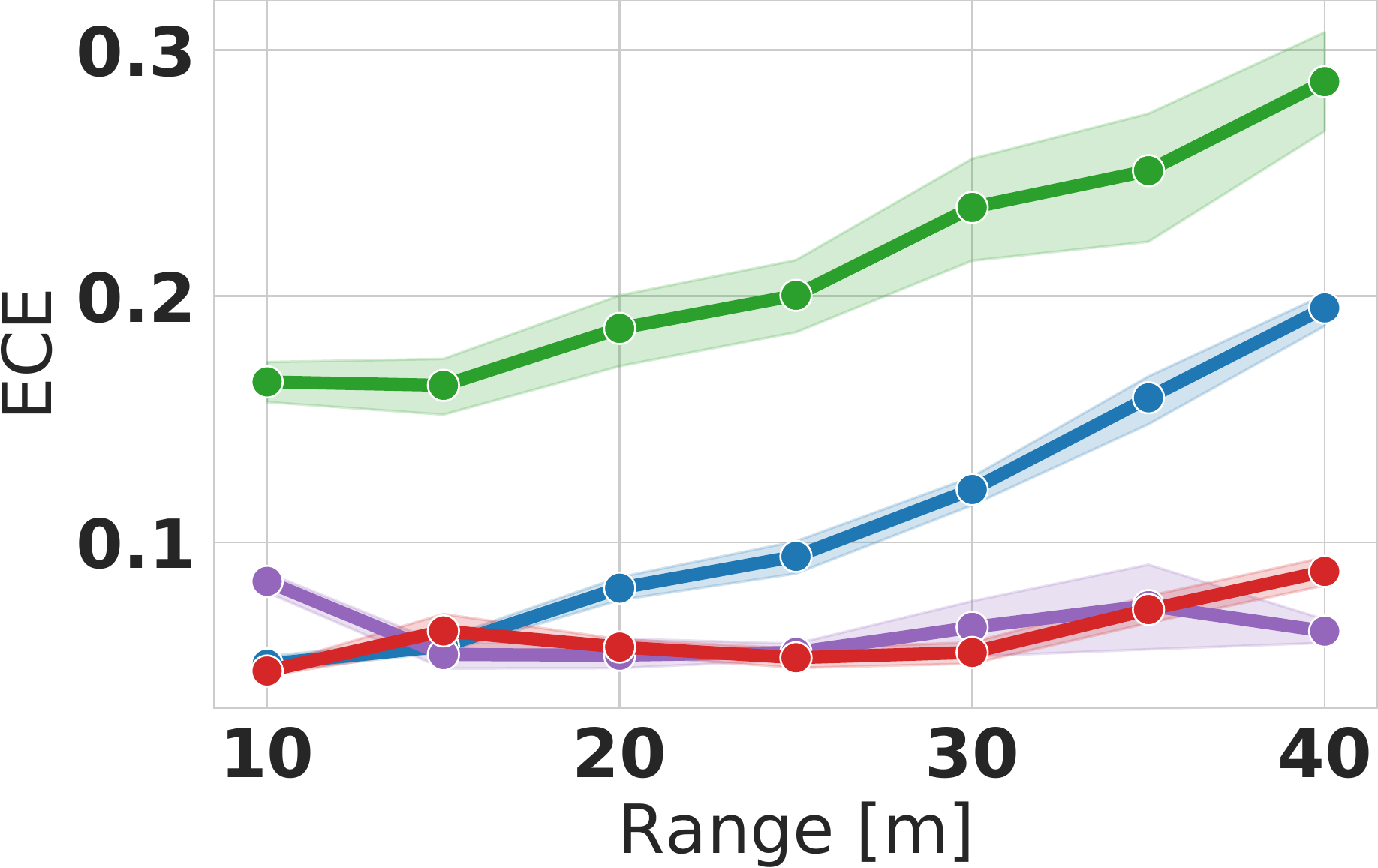}
\end{subfigure}
\hfill
\begin{subfigure}{0.48\columnwidth}
  \centering
  \includegraphics[width=1.0\textwidth, clip]{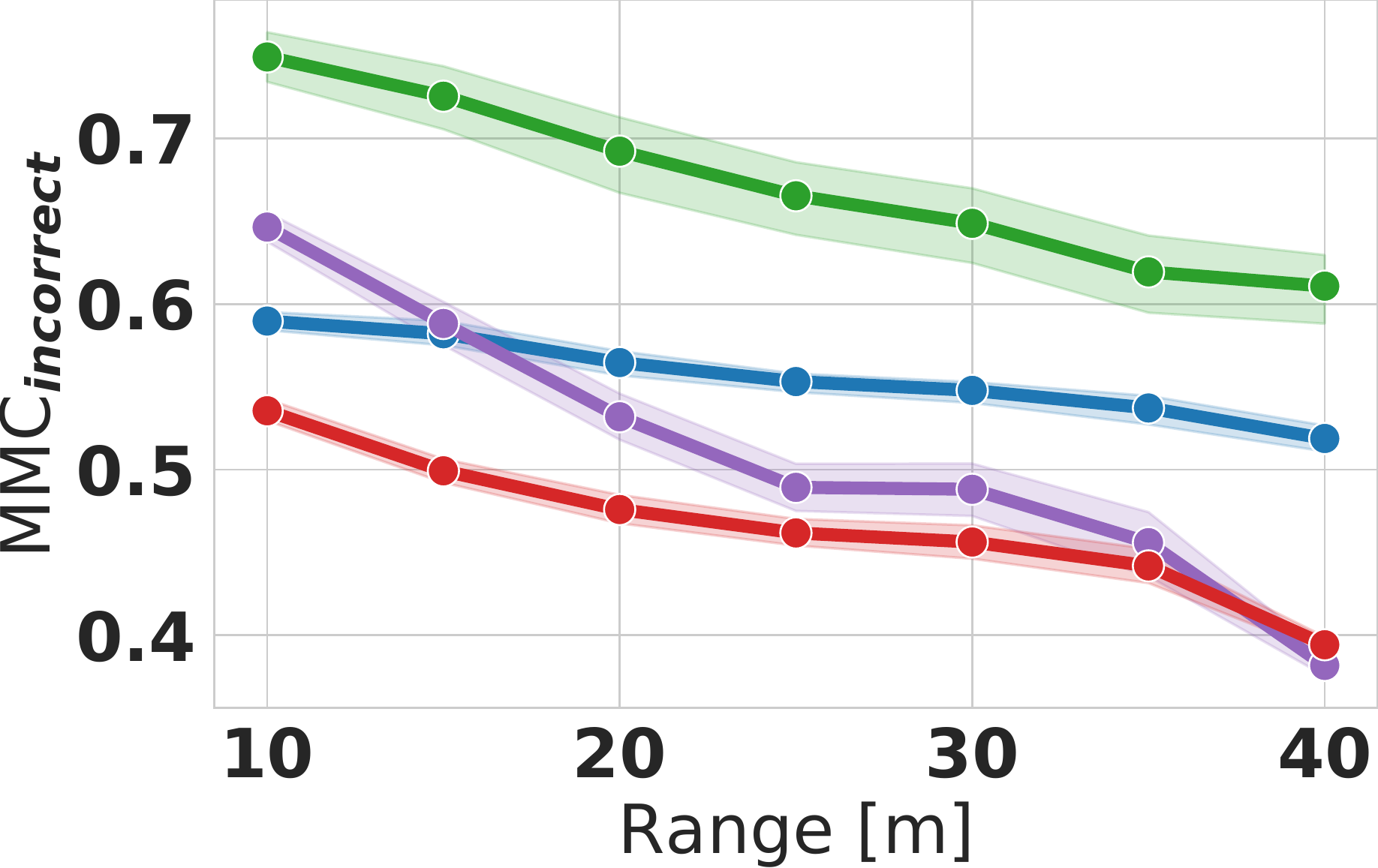}
\end{subfigure}
    \caption{Performance over range of all training methods on Env2-Test. 
    Creating subsets by grouping objects at similar distances to the sensor, we evaluate the ECE and \MMCneg.
    As expected, all methods perform worse at both metrics at larger distances with the best performance seen by R-/P-smoothing.
    }
\label{fig:performance_over_range}
\end{figure}

Both R- and P-smoothing were designed to improve the performance of the classifiers at larger range by lowering the confidences of the labels for these spectra.
We now study if this goal has been achieved.
In order to study the influence of the training methods over range, we create subsets, using $5$m thresholds from $10$m up to $40$m, of the test sets and evaluate the ECE and \MMCneg in Fig.~\ref{fig:performance_over_range}.

We find that using the smoothing methods which considers the uncertainty associated with the spectra to smoothen the hard labels, provide better calibrated outputs, especially at larger distances.
Unlike the Baseline and $\epsilon$-smoothing, which becomes significantly worse at ECE at larger distances, the calibration performance of R-/P-smoothing stay roughly constant at all ranges.
As depicted in Fig.~\ref{fig:teaser_baseline_vs_ours}, the accuracy of all classifiers become significantly worse at larger ranges, therefore it is crucial to lower the confidence of these distant samples.
We observe that R-/P-smoothing significantly reduce the confidences of the incorrect predictions at larger distances.
Whereas, the Baseline and $\epsilon$-smoothing still assign more than $50\%$ confidence on the incorrect predictions.
This result shows that the design choice of smoothing the labels of farther objects and low-power spectra significantly helps maintain calibrated predictions for uncertain samples which are predominantly found at larger ranges.

\section{Spectra corruptions}
\label{sec:p5_exps_spectra_corruptions}
\begin{figure} [!htp]
\centering
\begin{subfigure}{0.95\columnwidth}
  \centering
  \includegraphics[width=1.0\textwidth, clip]{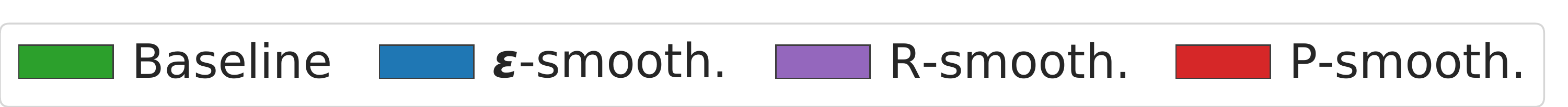}
  \label{legend}
\end{subfigure}
\begin{subfigure}{.48\columnwidth}
  \centering
  \includegraphics[width=0.95\textwidth, clip]{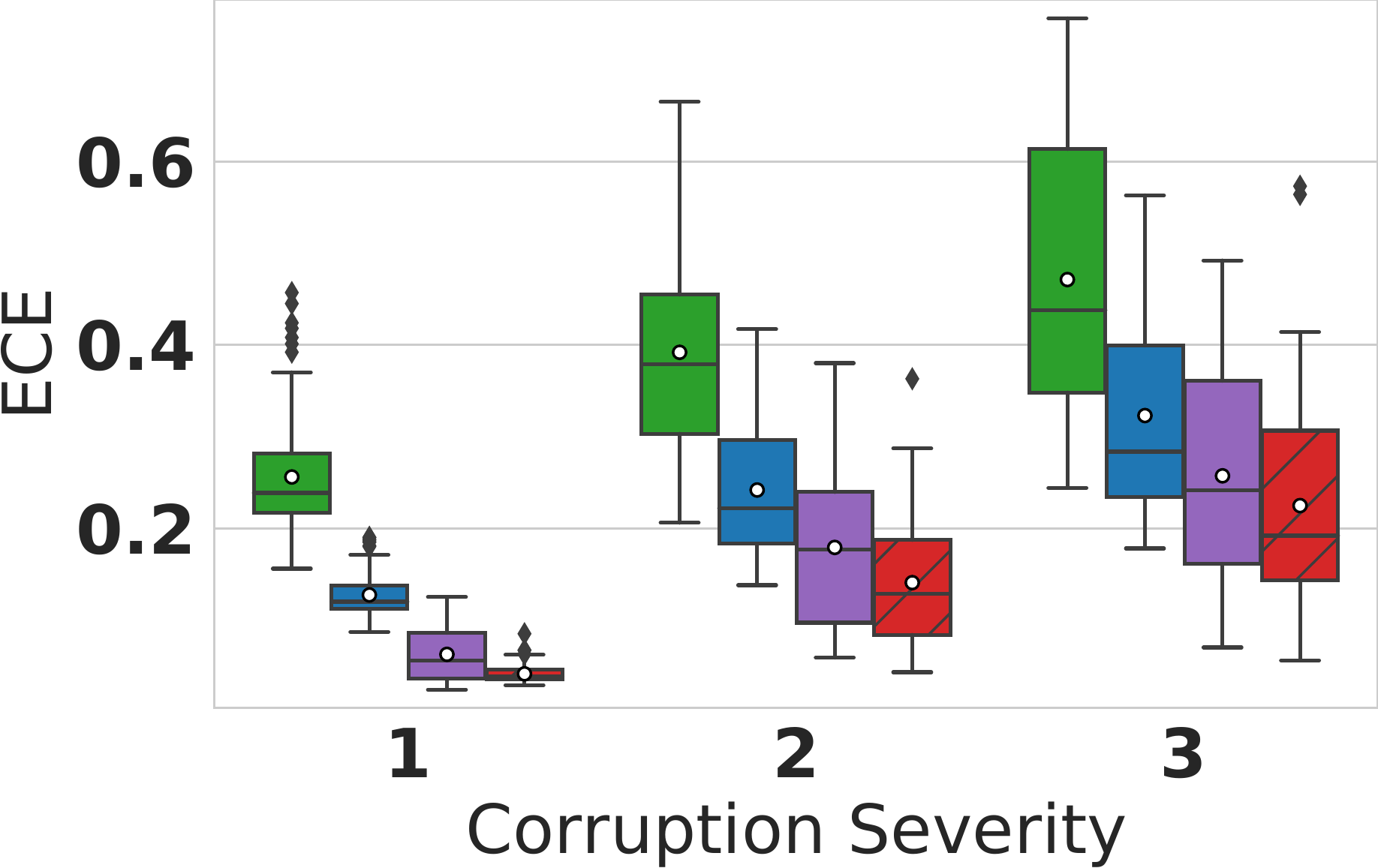}
  \caption{}
\end{subfigure}
\hfill
\begin{subfigure}{.48\columnwidth}
  \centering
  \includegraphics[width=0.95\textwidth, clip]{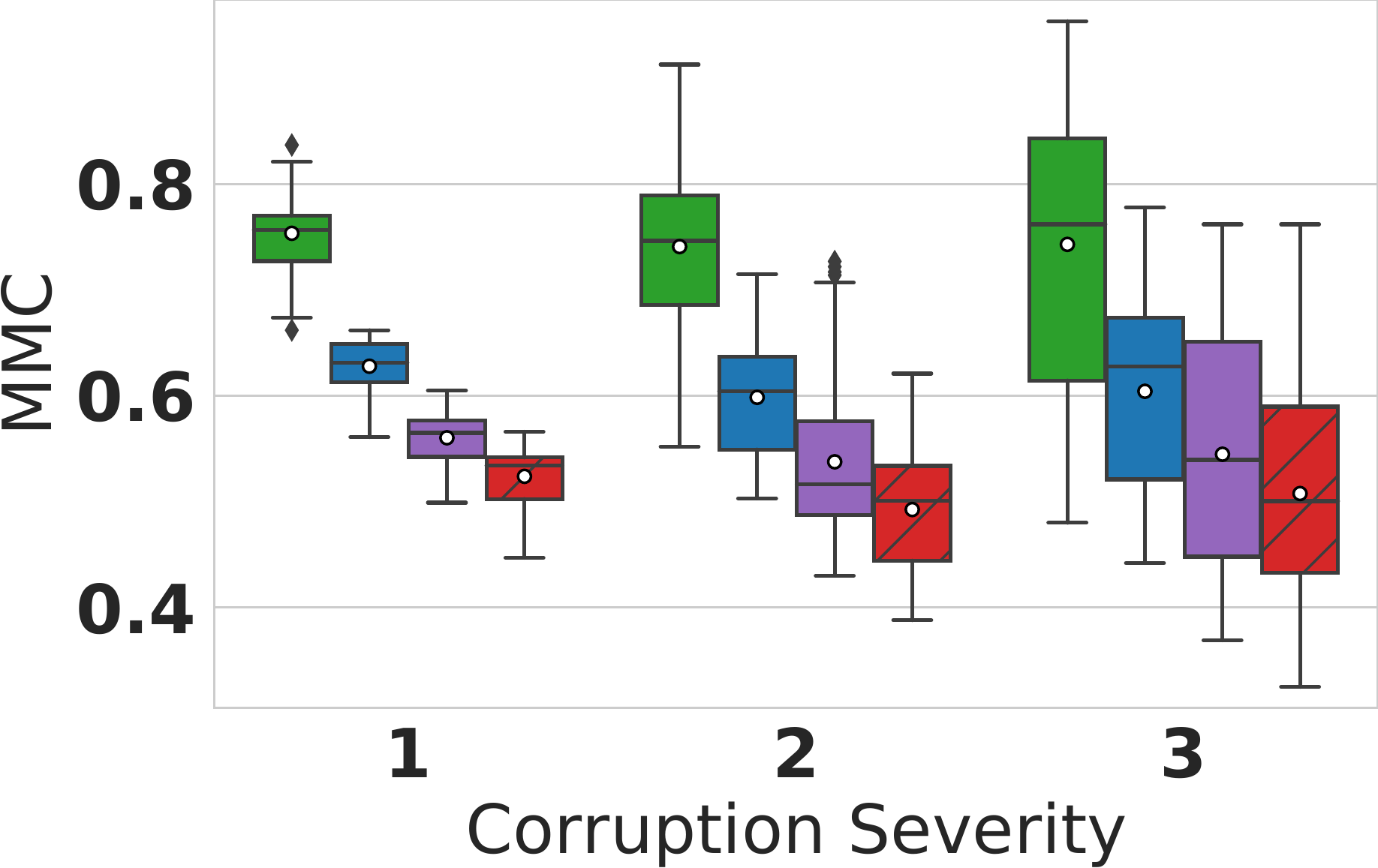}
  \caption{}
\end{subfigure}
    \caption{
        Boxplot visualizations of (a) ECE and (b) MMC evaluated on corrupted spectra samples from Env2-Test. 
    As expected, all ECEs become worse with higher severity and the label smoothing methods improve the calibration performance with best performance seen by P-smoothing.
    We also observe that the Baseline method keeps its average predicted confidences (i.e. MMC) roughly the same even though the classifiers are increasingly inaccurate and unfamiliar with the high severity corruptions.
    All corruptions have been averaged out per severity. 
    Hatched boxes indicate the best performance.
}
\label{fig:corruptions_performance_boxplots}
\end{figure}

In Fig.~\ref{fig:corruptions_performance_boxplots}, we study the ECE and MMC of the classifiers to unseen dataset shifts.
We use the same synthetic spectra corruptions introduced in~\cite{patel_radar21} to corrupt the test sets.
We average the metrics for all $7$ corruption types across the $3$ severities.
The Baseline accuracies at these $3$ severities are $50.1\%$, $33.2\%$ and $23.8\%$ .
As the corruptions become more severe, the ECE increases with the worst performance showed by the Baseline.
This is explained by the consistently high MMC of the Baseline as it fails to reduce its predicted confidences even though it becomes increasingly inaccurate with higher severities.
In summary, using soft labels greatly helps remedy the over-confidence on mis-classified samples, which becomes significantly worse under dataset distribution shifts.
The best performance is again observed by P-smoothing.

\section{Conclusion}
In an attempt to improve the calibration performance of radar classifiers, we proposed to incorporate label smoothing during training.
We developed two new techniques for producing soft labels using the range R and average received power P, which correlate with the underlying uncertainty inherent in the spectra.
This was the first work aiming to improve the predictive uncertainty quality for radar classifiers during the training process.
We find that the presented label smoothing methods achieve this goal and greatly improve the calibration performance of the classifiers.
In addition, as a side effect of addressing the mis-calibration and over-confidence issues of deep learning classifiers, we also observe consistent generalization performance gains.
As a result, R-/P-smoothing both produced more reliable and accurate classifiers, enabling better integration of such deep learning classifiers into real-world systems that rely on the predictive uncertainties to perform downstream actions. 
This work has shown that deep learning classifiers, which have the power of learning complex features from data, highly benefit from radar-specific knowledge and that further improvements can be made by tailoring learning algorithms to leverage this knowledge.
A future research directions for improving the performance of the radar classifiers is intervening during the training epochs to improve the uncertainty and generalization performance by incorporating even more radar specific knowledge.
The classifiers tend to focus on learning easy discriminative features in the data; thus, developing new ways to guide the model capacity to learn more complex and challenging features seems to be a promising direction.

\bibliographystyle{IEEEtran}
\bibliography{references}

\end{document}